\documentclass[10pt,journal,compsoc,anonymous]{IEEEtran}
\usepackage{amsmath,amssymb,amsfonts}
\usepackage{algorithmic}
\usepackage{textcomp}
\usepackage[ruled,linesnumbered]{algorithm2e}
\usepackage{bm}
\usepackage{graphicx}
\usepackage{booktabs}
\usepackage{stfloats}
\usepackage{enumitem}
\usepackage{xcolor}
\usepackage{multirow}
\usepackage{ulem}
\setitemize[1]{leftmargin=26pt}

\usepackage{amsmath}
\usepackage{CJK}

\begin{document}

\title{Dynamic Fair Federated Learning Based on Reinforcement Learning}

\author{Weikang Chen, 
        Junping Du*,
        Yingxia Shao,
        Jia Wang, 
        and Yangxi Zhou
\IEEEcompsocitemizethanks{\IEEEcompsocthanksitem Weikang Chen, Junping Du, Yingxia Shao, Jia Wang, and Yangxi Zhou are with the School of Computer Science (National Pilot School of Software Engineering), Beijing University of Posts and Telecommunications, Beijing Key Laboratory of Intelligent Telecommunication Software and Multimedia, Beijing 100876, PR China. 
E-mail: weikangchen@bupt.edu.cn, junpingdu@126.com, shaoyx@bupt.edu.cn, wangi2021110865@bupt.edu.cn, zhouyx@bupt.edu.cn.
}

\thanks{*Corresponding author: Junping Du (junpingdu@126.com).}
\thanks{This work was supported by the Program of the National Natural Science Foundation of China (62192784, U22B2038, 62172056).}
}

\markboth{The 5-th International Conference on Data-driven Optimization of Complex Systems}%
{Chen \MakeLowercase{\textit{et al.}}: Dynamic Fair Federated Learning Based on Reinforcement Learning}

\IEEEtitleabstractindextext{%
\begin{abstract}
Federated learning enables a collaborative training and optimization of global models among a group of devices without sharing local data samples. However, the heterogeneity of data in federated learning can lead to unfair representation of the global model across different devices. To address the fairness issue in federated learning, we propose a dynamic \textit{q} fairness federated learning algorithm with reinforcement learning, called DQFFL. DQFFL aims to mitigate the discrepancies in device aggregation and enhance the fairness of treatment for all groups involved in federated learning. 
To quantify fairness, DQFFL leverages the performance of the global federated model on each device and incorporates $\alpha$-fairness to transform the preservation of fairness during federated aggregation into the distribution of client weights in the aggregation process.
Considering the sensitivity of parameters in measuring fairness, we propose to utilize reinforcement learning for dynamic parameters during aggregation. 
Experimental results demonstrate that our DQFFL outperforms the state-of-the-art methods in terms of overall performance, fairness and convergence speed.
\end{abstract}

}

\maketitle

\IEEEdisplaynontitleabstractindextext

\IEEEpeerreviewmaketitle

\IEEEraisesectionheading{\section{Introduction}\label{sec:introduction}}


\IEEEPARstart{A}{s} an emerging sub-field of distributed optimization, federated learning is a learning approach that leverages edge clients with large communication capabilities and limited computational power to process data collection and model training \cite{li2020federated, li2022federated, zang2023fedpcf, guan2021federated, kou2018hashtag, meng2013tracking}. Unlike traditional distributed optimization, where privacy protection is paramount and local data on edge clients cannot be shared globally, FedAvg \cite{mcmahan2017communication} adopts a different approach. It transmits global models from the central server to the clients during each communication round between the central server and the edge clients and utilizes the local data on the clients to train the models. In each communication round, a subset of clients involved in the computation performs multiple local updates before these models are aggregated, enabling the learning of a global federated model using private data on individual clients while preserving privacy \cite{mo2022multi, li2018neural}.


In recent years, fairness has garnered significant attention in the context of federated learning \cite{li2019fair,cao2013review,meng2015high,li2018resilient,li2017tobit}. However, traditional federated learning algorithms often overlook the issue of fairness \cite{bonawitz2019towards,meng2015robust}, and federated aggregation models may exhibit potential biases towards certain clients or data labels \cite{li2023multi,li2022predicting,li2021heterogeneous}. For instance, although the global federated model may achieve high average accuracy on a test set, there is no guarantee of its accuracy on local data from individual clients \cite{wang2021novel}. This unfairness is exacerbated by the heterogeneous and inconsistent data sizes across clients in a federated learning environment, leading to significant variations in model performance among clients \cite{horvath2021fjord}.

Therefore, it is crucial to adopt fairness-based federated learning algorithms to address these disparities among individual clients and to design effective federated optimization methods that promote fairer performance of the global model without compromising its accuracy \cite{zhou2022chinese,xiao2022lecf,
shao2021memory,li2020distributed}. By incorporating fairness considerations into the federated learning framework, we can better align the interests of all parties involved and create a more equitable and trustworthy ecosystem for data collaboration and model training.

Due to privacy protection issues, direct access to raw data on each client and data distribution analysis on the client are not feasible \cite{lee2022privacy, li2022scientific, wei2019boosting, li2019application}. To address this challenge, this paper proposes the Dynamic Q Fairness Federated Learning Algorithm with Reinforcement Learning (DQFFL), drawing inspiration from q-FFL \cite{li2019fair} and PG-FFL \cite{sun2022fair}. The main objective of DQFFL is to alleviate differences in client-side aggregation and achieve fair treatment for all groups, including minority groups, in the global federated model.

The proposed algorithm transforms the maintenance of fairness during federated aggregation into the assignment of client weights to federated aggregates. It incorporates $\alpha$-fairness \cite{jung2019fairness} to achieve a balance between optimal model performance and fairness. Considering the sensitivity of fairness measurement parameters to data, DQFFL dynamically adjusts parameters by optimizing the aggregated weight-solving model and introducing reinforcement learning algorithms.

Compared to q-FFL, DQFFL avoids manual adjustments of q values by employing a policy update algorithm to automatically determine the optimal q values for the current federated communication round. Compared to PG-FFL, DQFFL takes into account reducing client computational costs and simplifying the action space complexity of intelligence, thereby accelerating convergence to the optimal policy.

The contributions of this paper can be summarized as follows:

\begin{enumerate}

\item We propose a reinforcement learning-based fair federated learning algorithm (DQFFL) to optimize the federated aggregation process for fair federated learning. DQFFL analyzes the training loss of the federated model at each client participating in the computation to serve as a fairness measure. This approach is privacy-preserving and incurs only a small communication and computation overhead.


\item We design a dynamic adjustment mechanism based on reinforcement learning for fairness. The training loss and accuracy of each client in federated learning were used as the state of an agent. The federated aggregation weights is dynamically adjusted based on the output actions of the agent, allowing for the dynamic quantification of fairness during the federated learning process.


\item We introduce a parameter sensitivity strategy based on $\alpha$-fairness. By analyzing the fairness metric of $\alpha$-fairness and the training method of federated learning, we adopte an $\alpha$-fairness variant to reduce the dependence of parameters on the dataset. This approach not only reduces the cost of optimal reinforcement learning strategies but also accelerates model convergence.


\item Experimental results on different datasets show that DQFFL can effectively address the challenge of balancing fairness and effectiveness in the federated learning process. The performance of DQFFL outperforms other schemes, demonstrating its superiority in achieving fair and effective federated learning.

\end{enumerate}
 
\section{RELATED WORK}

Classical federated learning algorithms aggregate models from different clients based on the size of their local training data, leading to a global federated model \cite{jung2019fairness,li2017variance}. However, the heterogeneity of data size and distribution across clients can disproportionately affect some clients' model performance, resulting in inconsistent results \cite{li2021ditto}. Despite a high average accuracy on the test set, the global federated model may not be accurate on individual devices. Research has been focusing on addressing fairness in federated learning with heterogeneous data \cite{sultana2022eiffel}.


Some researchers have focused on addressing fairness issues during federated aggregation by analyzing gradient conflicts returned by clients \cite{mitra2021linear}. AFL employs a minimum value optimization approach to ensure overall fairness by prioritizing optimization for the worst-performing clients without compromising others. FedFVG \cite{wang2021federated} uses cosine similarity to detect and eliminate gradient conflicts, thereby reducing potential unfairness among clients during aggregation. However, selecting clients for optimization in each round of federated communication may lead to larger gradient variance, possibly causing convergence stability problems.

Another approach involves using fairness metrics such as Jain's Index and Gini coefficient to quantify fairness in federated aggregation based on loss and accuracy metrics from individual clients \cite{cotter2019optimization}. The aim is to adjust aggregation weights from a fair allocation perspective to achieve fairness during federated learning. For instance, q-FFL adopts resource allocation principles by assigning larger weights to clients with higher loss values during training, promoting a balanced distribution during federated aggregation. It also proposes q-FedAvg, a FedAvg-like approach, and employs the estimated Lipschitz constant to avoid adjusting the learning rate for different q values. On the other hand, $\alpha$-FedAvg \cite{Jia2022afair} incorporates Jain's Index and $\alpha$-fairness calculations to achieve fairness in the federated model system. PG-FFL uses the Gini coefficient to measure the global federated model's performance on each client and employs reinforcement learning to optimize model accuracy and fairness \cite{zhang2021deep}. DRFL \cite{zhao2022dynamic} combines $\alpha$-fairness with loss bias and client selection policies to dynamically assign weights to each client, promoting fairness in federated learning. However, using $\alpha$-fairness metrics may require parameter tuning depending on the dataset and federated environment, and such methods lack dynamic adaptability as the metrics are calculated in a predetermined manner during each round of federated learning.


The proposed Dynamic Q Fairness Federated Learning algorithm (DQFFL) addresses the aforementioned challenges. It achieves fairness in the federated learning model by dynamically balancing the reinforcement policy update algorithm, taking into account the computational cost of clients to avoid gradient adjustments based on client returns. The loss calculation of each client during each round of communication serves as the fairness state, and the algorithm intelligently optimizes the aggregation weights by taking appropriate actions based on the current situation, effectively achieving fairness adjustment in federated learning.

\subsection{Federated Learning}

In the standard federated learning framework, there are K clients, denoted as $k$, each with a local dataset of size $s_k$. These clients update the model using their respective local datasets and send the updated model to a central server for aggregation. The goal is to find the global federated model parameter vector $w$ that minimizes the empirical risk objective:
\begin{equation}
    \mathop {\min }\limits_\omega  {\mkern 1mu} F(\omega ) = \sum\limits_{k = 1}^K {{p_k}{F_k}(\omega )}  = \sum\limits_{k = 1}^K {\frac{{{s_k}}}{{\sum\limits_{i = 1}^K {{s_i}} }}{F_k}(\omega )} 
\end{equation}
where $p_i$ denotes the weight value that the model obtained from client-side training contributes to the global model. The local target $F_k$ can be defined by the empirical risk on the local data. The standard algorithm for solving (1) is the Federated Average Aggregation algorithm\cite{yang2021characterizing}, FedAvg is executed in communication rounds, where in each communication round, the central server selects $m=\max{CK, 1}$ clients participating in this round according to a certain probability, and C denotes the probability value. The selected clients receive the global model\cite{blum2021one} from the central server, update the global model using the local data on the clients, and send the results back to the central server after completing the model update. The central server, after all the participating clients have completed the $p_k=s_k \sum^m{i=1}s_i$ update the global parameter return, the central server according to the probability distribution model. 

Although FedAvg has a high communication efficiency and achieves high overall accuracy in $p_k$ experiments, the selection with unbiased weights, Client and server aggregation can lead to fairness issues \cite{agarwal2018reductions}. Specifically, clients with larger dataset size $s_k$ are more frequently and better optimized than clients with smaller $s_k$ \cite{spiridonoff2021communication}, which typically represent a small number or outliers because they are less likely to be treated fairly due to the small percentage value of data size on the local client.

\subsection{Fairness theory}
Fairness in machine learning is usually defined as the protection of some specific attributes. In the resource allocation problem, decision makers need to consider the fairness of resource allocation to satisfy the overall goal of optimality \cite{kim2022adaptive}. $\alpha$-fairness is a widely used fairness indicator in the resource allocation problem:
\begin{equation}
{U_\alpha }(x) = \left\{ {\begin{array}{*{20}{c}}{\ln (x),}&{\alpha  = 1}\\{\frac{1}{{1 - \alpha }}{x^{1 - \alpha }},}&{\alpha  \ge 0,\alpha  \ne 1}\end{array}} \right.
\end{equation}
where $U_{\alpha} (x)$ denotes the benefit of some specific user given x allocated resources, and the goal is to find a resource allocation strategy that maximizes the sum of benefits for each user. Based on the above definition and quantifiers of fairness in resource allocation and combined with the characteristics of federated learning, the global model for solving federated learning can be considered as a resource allocation aimed at serving users.

\begin{figure*}
  \centering 
  \includegraphics[scale=0.5]{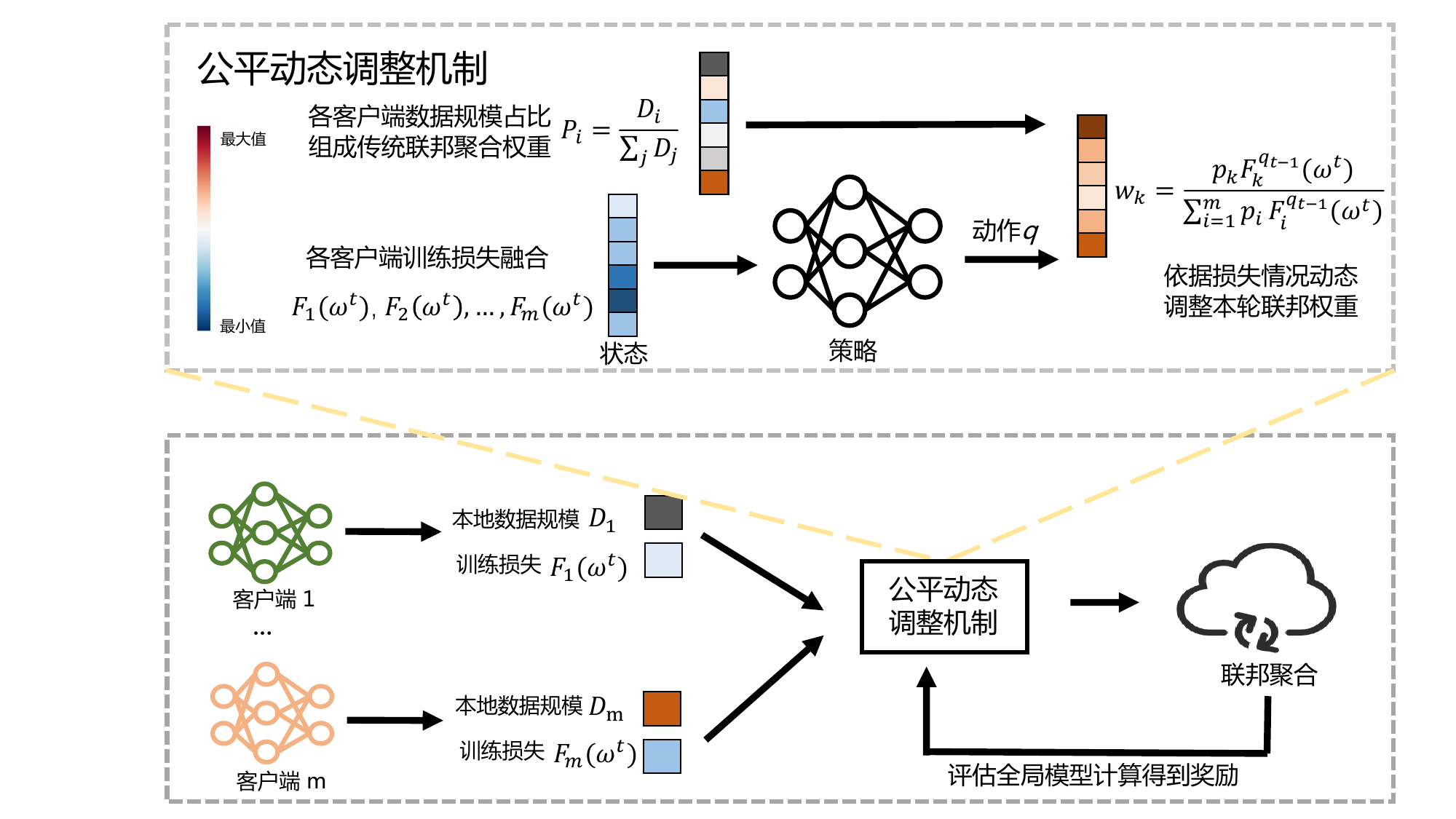}
  \caption{Illustration of dynamic fair reinforcement federated learning algorithm.The whole framework diagram is divided into two main parts from the reinforcement learning perspective: The upper side shows the fair dynamic adjustment mechanism, and the lower side shows the federated learning process. From the perspective of federated learning, the detailed process of fair reinforcement learning algorithm is divided into 3 Steps.}
  \label{fig1}
\end{figure*}

\section{METHODOLOGY}
In this section, this paper first gives the dynamic fair reinforcement federated learning algorithm proposed in this paper. Then, this paper designs the framework of fair reinforcement federated learning model based on policy update. Finally, this paper compares the communication computational overhead of the algorithm proposed in this paper and the existing algorithms, and analyzes the advantages and disadvantages of the algorithm proposed in this paper.
\subsection{Fair Reinforcement federated Learning Algorithm}
This section of the paper gives the dynamic fair reinforcement federated learning algorithm:
\begin{equation}
F({\omega ^{t + 1}}) = \sum\limits_{k = 1}^K {\frac{{{p_k}F_k^{{q_{t - 1}}}({\omega ^{t - 1}})}}{{\sum\limits_{i = 1}^K {{p_i}F_i^{{q_{t - 1}}}({\omega ^{t - 1}})} }}} {F_k}({\omega ^t})
\end{equation}
where $w_t$ denotes the final global federated model obtained at the $t$ round, and $w^{t-1}$ denotes the federated model obtained at the previous round of federated communication computation; also the parameter $q_{t-1}$ is time-dependent and is obtained from the actions made by the intelligence in the reinforcement learning algorithm according to the losses as states of the participating training clients, realizing that in the process of federated learning is dynamically adjusted. The client $k$ weights $w_k$ for the federated aggregation of this algorithm are calculated as
\begin{equation}
{w_k} = \frac{{{p_k}F_k^{{q_{t - 1}}}({\omega ^t})}} {\sum\limits_{i = 1}^m {{p_i}F_i^{{q_{t - 1}}}({\omega ^t})} }
\end{equation}
when $q=0$, this dynamic fairness reinforcement federated learning algorithm is equivalent to the FedAvg algorithm; when $q=1$, the weight value of the current federated aggregation is equivalent to the performance of the previous round of the federated model on the participating clients. That is, the fairness to each client at the current federated aggregation is dynamically adjusted during each training round using the losses of the participating clients in the previous round, with higher loss clients being assigned larger weights and the opposite for lower loss clients.

Compared with the uncertainty of the q-FFL algorithm on the range of hyperparameter $q$ values, this algorithm uses the loss of each participating client from the previous round of training and the fused $\alpha$-fairness formula to weaken the sensitivity to the range of parameter $q$ values in the form of a ratio, which further limits the action space of the intelligence and facilitates the convergence of reinforcement learning. And in the training process of q-FFL algorithm, in addition to the gradient calculation, q-FFL also needs to calculate the model parameter difference and Lipschitz Gradient at the client side, which leads to potential aggregation information leakage and is not conducive to system security; considering that the difference of loss values on each client decreases as the training proceeds in the federated learning phase, this algorithm in DRFL algorithm based on the determination of the parameter q values is defined as a Markov decision process, where the state is represented by the locally calculated loss of the model for each client in each round. Also, the federated aggregation training is completed by using one model downlink and upload in each round of federated communication, which is more stable and efficient compared to DRFL that requires two message passes in one round of federated training.

\subsection{Reinforcement Learning in Fair federated Learning}
In federated learning training, the optimal weight distribution is non-differentiable because a certain percentage of clients are randomly selected to participate in the update in each round and the performance of the model on each client needs to be considered during aggregation. In this paper, the problem of weight distribution of different local models in the global model is modeled as a deep reinforcement learning problem in this paper to solve the problem of considering client accuracy and fairness balance in the federated aggregation stage. Reinforcement learning differs from supervised learning in that it does not rely on feedback information given by labels and does not require explicit correction of sub-optimal operations. 

The focus of reinforcement learning is to find a balance between exploring the unknown domain and developing current knowledge \cite{kim2022adaptive}. When the intelligence running on the server follows the policy $\pi$ based on the losses returned by the client as the state s and performs the action a as the dynamic q value, the probability of a T step trajectory is obtained as 
\begin{equation}
P(\tau \left| \pi  \right.) = {\rho _0}({s_0})\prod\limits_{t = 0}^{T - 1} {P({s_{t + 1}}\left| {{s_t},{a_t}} \right.)\pi (\left. {{a_t}} \right|{s_t})}
\end{equation}
where $s_0$ denotes the initial state and $\rho _0$ denotes the initial state distribution; $P(s_{t+1}|s_t, a_t)$ denotes the probability of shifting from state st and the state to $s_{t+1}$ after the intelligence performs action at according to the policy. The reward function $R(\tau)$ is used in reinforcement learning to evaluate the merit of an intelligence's strategy. The objective of deep reinforcement learning is to find an optimal strategy $\pi$ that maximizes the long-term reward expectation:
\begin{equation}
{\pi ^*} = \mathop {\arg \max }\limits_\pi  J(\pi )
\end{equation}
Among the long-term incentive expectations:
\begin{equation}
J(\pi ) = \int_\tau  {P(\tau |\pi )R(\tau )}  = \mathop E\limits_{\tau  \sim \pi } [R(\tau )]
\end{equation}
Combining the federated learning training process with the need to change state after each round of communication and the need to rely on reinforcement learning to set weights for their next round of federated aggregation \cite{zhang2021adaptive},  in this paper, we use strategy optimization to train the intelligence and update the strategies using gradient optimization by:
\begin{equation}
{\nabla _\theta }J({\pi _\theta }) = \mathop E\limits_{\tau  \sim {\pi _0}} [\sum\limits_{t = 0}^T {{\nabla _\theta }\log {\pi _\theta }({a_t}\left| {{s_t}} \right.)R(\tau )}]
\end{equation}
In the algorithm proposed in this paper, the design requirements of reinforcement learning for state, action and reward are met by combining federated learning with each round of communication turnover:
\begin{enumerate}
\item State: The loss of the global federated model on the client in the $t$ round of federated communication constitutes the state vector $S_t=\{F_1(w^t), F_2(w^t), ..., F_k(w^t)\}$, The server-side and client-side form an environment jointly maintained during the training process, and the global federated model in The model $w^t$ is updated for each round of communication , and the loss value $F(w^t)$ is calculated using the current model on the clients involved in the computation.
\item Action: In each round of federated communication, when the state is updated, the intelligent body needs to perform the corresponding action according to the current state, i.e., to determine the current parameter q value based on the loss situation calculated by the current participating training clients, and to realize the fairness between each participating computing client during the federated training process by dynamically evaluating the aggregation.
\item Reward: The reward is then used to balance fairness and accuracy in the federated learning process and to optimize the federated global model to achieve the best results. Defined as:
\begin{equation}
{r^{t - 1}} = {a^t}{e^{ - {\mathop{\rm var}} ({F_1}({\omega ^t}),{F_2}({\omega ^t}), \cdots ,{F_k}({\omega ^t}))}}
\end{equation}
where $a^t$ indicates the performance of the current model on the test set, and such settings encourage the joint model to achieve optimal and fair performance.
\end{enumerate}

\begin{table*}
\centering
\caption{Comparison of Communication and Computational Overhead of Different Federated Learning Algorithms}
\label{tab1}
\begin{tabular}
{|c|c|c|c|c|}
\hline
Algorithm & Server Communication & Client Communication & \begin{tabular}[c]{@{}c@{}}Calculation Complexity\\  in Server\end{tabular} & \begin{tabular}[c]{@{}c@{}}Calculation Complexity\\  in Client\end{tabular} \\ \hline
FedAvg           & O(mP)      & O(P)      & O(1)         & O(P)     \\ \hline
$\alpha$-FedAvg  & O(mP)      & O(P)      & O(1)         & O(P+A)   \\ \hline
q-FFL            & O(m(P+e))  & O(P+e)    & O(1)         & O(P+e)   \\ \hline
DRFL             & O(2mP)     & O(2P)     & O(1)         & O(P)     \\ \hline
PG-FFL           & O(m(P+e))  & O(P+e)    & O(1+RL)      & O(P+A)   \\ \hline
DQFFL            & O(m(P+e))  & O(P+e)    & O(1+RL)      & O(P)     \\ \hline
\end{tabular}
\end{table*}

\subsection{Fair Reinforcement federated Learning Algorithm Process Framework}
Synthesizing the above, the flow of the fair reinforcement federated learning algorithm proposed in this paper is summarized as Algorithm 1. The executors in Algorithm 1 are divided into a server and a client. The server is responsible for selecting clients to participate in the training of federated learning and maintaining the optimization of the reinforcement learning intelligence' strategies to achieve fair aggregation, and the reinforcement learning uses the current loss of the global model of the federated on each client as the state to perform the corresponding action according to the current optimal strategy, and the server uses the output of the intelligence' actions as the weight of each client's model in the completion of the federated aggregation with the fairness q value to achieve a higher penalty for clients with high loss clients are assigned greater penalties, adjusting for smaller values in the aggregation phase due to smaller data size. The client is responsible for updating the global federated model based on local data and using the performance of the global model on local data to provide the basis for the server's fairness metric. This algorithm ensures that after the introduction of reinforcement learning, the best policy obtained through reinforcement learning optimizes the trade-off between effectiveness and fairness in the federated learning process while ensuring data privacy, so that the final federated global model has a more balanced performance on local data on each client.

Figure 1 illustrates the fair reinforcement federated learning framework proposed in this paper. The whole framework diagram is divided into two main parts from the reinforcement learning perspective: The upper side shows the fair dynamic adjustment mechanism, and the lower side shows the federated learning process. From the perspective of federated learning, the detailed process of fair reinforcement learning algorithm is divided into 3 Steps:
\begin{enumerate}
    \item The server first determines the clients participating in this training by random selection, and broadcasts the training parameters of the global federated model $w^t$ for the current round to these clients, which will adopt the global federated model $w^t$ instead of the local model.
    
    \item The client selects a batch size $B$ of data to compute the loss $F(w^t)$ using the model $w^t$ ; and updates the model $w^t$ using a local optimization algorithm on the training set; after the computation is completed, the loss is sent to the server together with the model.

    \item In the current round, the server receives the computation results of all the clients participating in this round, and first inputs the loss as the state to the intelligence, which gives the corresponding action as the q value. According to Equation 5, the server completes the federated aggregation calculation based on the product of the client's data scale share and the $q$ power of its loss as the federated aggregation weight.
\end{enumerate}
\subsection{Communication overhead and computational overhead}
Table 1 shows the comparison of communication overhead and computation overhead between different algorithms in each round of federated aggregation, where the participation in training is noted as m clients. The size of the model parameters to be trained, P, and the size of the additional data parameters are all e, $e < P$ . A denotes the overhead of predicting the local test set.

\textbf{Communication overhead}. The communication overhead is calculated by considering the number of message interactions and the size of the data carried in each round of federated communication. FedAvg first randomly selects the clients participating in this round of computation in each round of federated communication and distributes the global federated model to each client; when the local training is completed, the training parameters of the model are transmitted back to the server, and this communication overhead is denoted as $O(mP)$ in this paper; the communication overhead of $\alpha$-FedAvg is also consistent with FedAvg. communication overhead is also consistent with FedAvg. q-FedSGD and q-FedAvg implemented by the q-FFL algorithm require an additional $h^t_k$  to be passed to the server when passing back to the server, compared to the FedAvg algorithm. The communication overhead of q-FFL is therefore recorded as $O(m*(P+e))$; the DRFL algorithm is different from FedAvg's random selection strategy, which requires two communication steps to determine the clients participating in training in this round, and therefore its communication overhead is recorded as $O(2mP)$; PG-FFL is similar to q-FFL in that it requires an additional The communication overhead is $O(m*(P+e))$; the algorithm proposed in this paper requires an additional loss function in addition to the model training parameters for each round back to the server. All clients only generate communication with the server, and only DRFL requires two message interactions in each round of communication; the difference in 
communication overhead between the clients of each algorithm is the different amount of data returned in each round of communication.

\textbf{Computational overhead}. The computational overhead of the server is $O(1+RL)$, except for PG-FFL and the algorithm proposed in this paper, which require an additional update of the reinforcement learning intelligence policy, denoted as $O(1+RL)$, and the rest of the algorithms complete the federated aggregation with O(1) computational complexity. As for the client-side overhead, $\alpha$-FedAvg and PG-FFL require additional client-side completion of the accuracy computation of the federated model on the local test dataset compared to FedAvg, denoted as $O(P+A)$; q-FFL algorithm. Compared to the FedAvg algorithm, which additionally needs to calculate ht  for weight calculation, which is written as $O(P+e)$; while the DRFL and the algorithm proposed in this paper need to calculate the local loss to obtain the aggregated weights, and the local loss is the intermediate result obtained when using the gradient update algorithm, so no additional computational overhead is introduced, but the calculations are all put on the server to reduce the client side of the data processing The computational pressure and overhead of data processing are reduced on the client side.

Based on the above analysis, the algorithm proposed in this paper integrates the performance of the client side and reduces the computational pressure on the client side; at the same time, it provides the training situation of the client side to the server with a smaller communication cost and the number of information exchanged in each round of communication, and dynamically adjusts the accuracy and fairness of the federated learning process with the reinforcement learning algorithm running on the server side under the premise of ensuring privacy.

\section{Experiment}
In this section, we evaluate the algorithms proposed in this paper. Firstly, we compare the fairness of the proposed algorithms and existing algorithms on multiple datasets. Then, we further analyze the impact of agents on federated aggregation to verify the effectiveness of the proposed algorithm.

\subsection{Experimental setup}
In this experiment, we tested a series of federated datasets using common models combined with the proposed federated learning algorithm. The selection and introduction of the q-FFL algorithm article were consistent. Table 2 summarizes the statistical data of the federated datasets, Synthetic \cite{shamir2014communication}, Vehicle \cite{smith2017federated} and Sent140 \cite{go2009twitter} used in the federated learning experiment. In this experiment, we randomly split each local device's data into 80\% training set, 10\% testing set, and 10\% validation set.

%
%

\begin{table}[]
\centering
\caption{Statistics of Federated Datasets}
\begin{tabular}{|c|c|c|}
\hline
Dataset   & Clients & Sample \\ \hline
Synthetic & 100     & 12697  \\ \hline
Vehicle   & 23      & 43695  \\ \hline
Sent140   & 1101    & 58170  \\ \hline
\end{tabular}
\end{table}

\subsection{Fairness comparison experimental results}
the q-FFL algorithm and $\alpha$-FedAvg algorithm are selected for comparison with the algorithm proposed in this paper. In order to exclude the influence of irrelevant factors, the number of clients involved in the computation in each round of aggregation is set to 10 in this paper. The learning rates of Synthetic, Vehicle, and Sent140 are 0.1, 0.01, and 0.03, respectively, and the batch sizes of Synthetic, Vehicle, and Sent140 are 10, 64, and 32. The number of local iterations $E$ is fixed to 1. The accuracy of the final model is tested on each participant's local dataset by conducting several sets of experiments with the above dataset, and the frequency distribution histogram is constructed by counting the customers whose values fall into the corresponding intervals.

\begin{table*}[!t]
\centering
\caption{Comparison of Accuracy Distribution of Different Federated Learning Algorithms}
\label{tab:my-table}
\begin{tabular}{|c|c|c|c|c|c|}
\hline
Dataset   & Algorithm  & Average\% & Worst 10\% & Best 10\% & Variance \\ \hline
 
& FedAvg   & 81.7     & 23.6      & 100.0    & 597      \\ \cline{2-6} 
Synthetic   & q-FFL(q=1) & 79.1     & 42.4    & 96.3     & 302      \\ \cline{2-6} 
& $\alpha$-FedAvg   & 82.1   & 40.2   & 100.0    & 341      \\ \cline{2-6} 
& DQFFL    & 80.1     & 43.5      & 94.7     & 331      \\ \hline

& FedAvg     & 85.9     & 47.3      & 95.2     & 247.3    \\ \cline{2-6} 
Vehicle & q-FFL(q=1) & 87.3     & 68.7      & 93.7     & 48.2     \\ \cline{2-6} 
& $\alpha$-FedAvg   & 87.1     & 69.9      & 94.0     & 40.6     \\ \cline{2-6} 
& DQFFL      & 87.5     & 73.5      & 94.1     & 34.8     \\ \hline

& FedAvg     & 69.4     & 35.4      & 95.1     & 345.7    \\ \cline{2-6} 
Sent140 & q-FFL(q=1) & 68.4     & 45.3      & 90.0     & 157.0    \\ \cline{2-6} 
& $\alpha$-FedAvg   & 70.0     & 43.1      & 92.9     & 197.1    \\ \cline{2-6} 
& DQFFL      & 70.2     & 45.5      & 93.2     & 176.5    \\ \hline
\end{tabular}
\end{table*}

Table 3 further shows the performance of each comparison algorithm on the dataset, including the average accuracy, the worst performing (10\%) in the category accuracy, the best performing (10\%) accuracy in the category, and the variance of the fairness metric used to measure the model. From the analysis of the experimental results, it is obtained that the algorithm DQFFL proposed in this paper achieves good performance before optimizing the model performance and fairness. The average accuracy of DQFFL on the Synthetic dataset rate decreases by 1.6\% compared to FedAvg, but reduces fairness by 40\% by adjusting the performance of the worst and best performing clients. From the Vehicle, Sent140 dataset performance, DQFFL compared to other comparison algorithms, the algorithm further reduces the variance by boosting the accuracy and average accuracy of the worst performing clients in the category at the expense of the accuracy of the best performing clients in the category, taking into account the performance of reinforcement federated learning on fairness in a comprehensive manner.

\subsection{Experiments to validate the effectiveness of the reinforcement learning module}
This section verifies the effectiveness of the reinforcement learning module for dynamic adjustment of fairness q-values. To simulate a federated environment, in this paper, a non-independent identically distributed sampling of the dataset FMNIST is performed, and a total of 60 clients are generated, each containing data and labels for all categories, while the model is used for classification using 2-layer convolutional layers and a multi-layer perceptron.\\
Table 4 demonstrates the performance of the final trained model on the dataset FMNIST with different values of fairness q-values adjusted for the federated aggregation stage. The q-values dynamically adjusted by DQFFL using reinforcement learning are similar in form to those of DR-FedAvg. From the experimental results, DQFFL outperforms DR-FedAvg in terms of accuracy and the average value of each label; the algorithm PGF-FedAvg algorithm uses reinforcement learning to adjust the weights in the form of Gini coefficients, and from the experimental results, the labels Shirt, Pullover is inferior to DQFFL.

\begin{figure}[htb]
	\centering
	\includegraphics[width=0.48\textwidth]{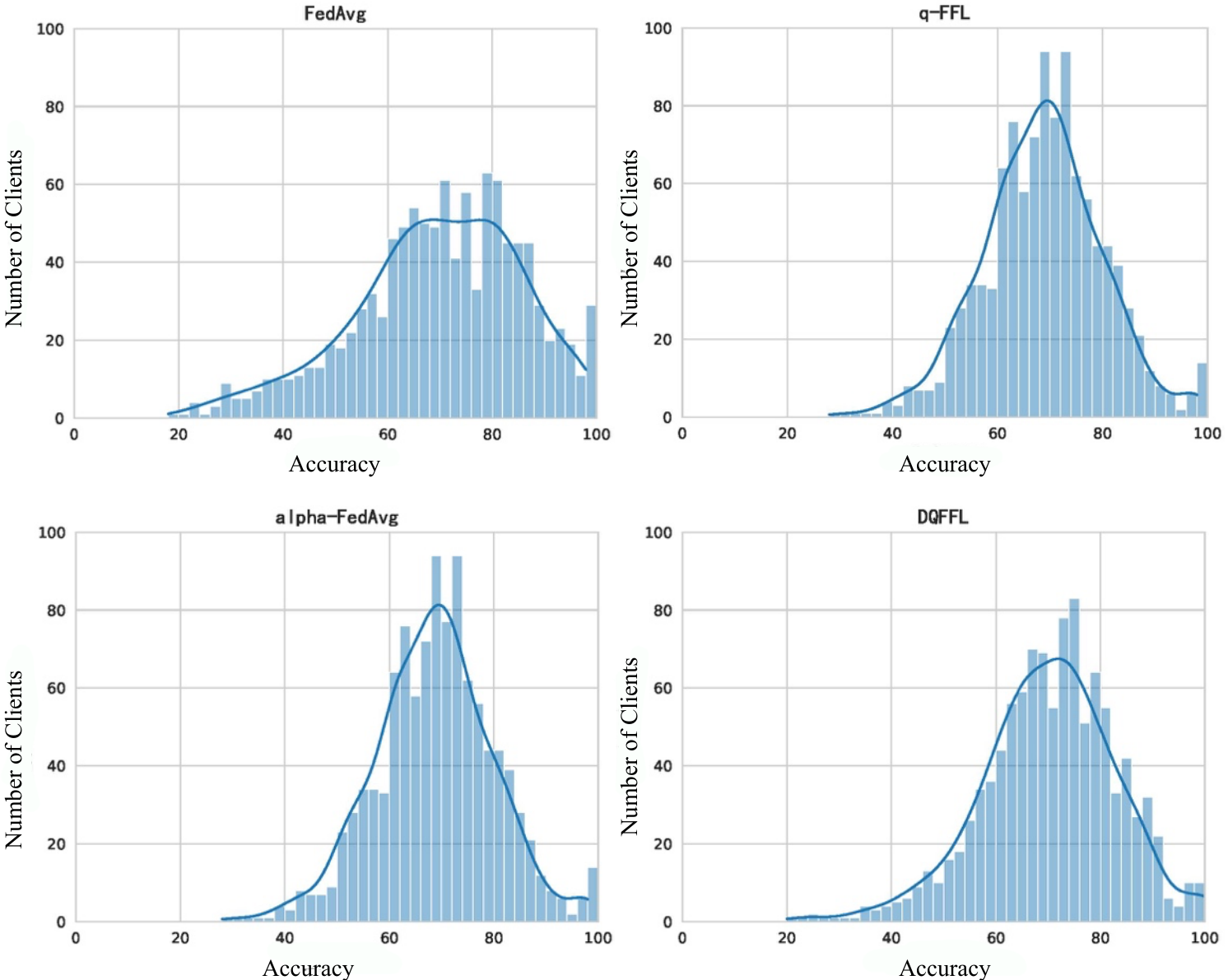}
	\caption{Histogram of Frequency Distribution of Each Algorithm on the Sent140 Dataset.}
\end{figure}

\subsection{Experiments to validate dynamic tuning of reinforcement learning modules}
In this section, we verify that the intelligent body can effectively and dynamically adjust the q-value even if the total number of clients is fixed at 100 clients but the number of clients participating in each round of federated training is changed, as shown in Figure 3. The number of clients participating in each training round is selected from 10, 30, 50, and 70, respectively, and the experimental results show that the output value of the intelligent body can effectively learn the q-value and dynamically adjust it according to different federated environments, which verifies that the reinforcement learning module can adapt to the impact of the change in the number of participating clients on the data distribution.
\begin{table}[]
    \centering
    \caption{Validation Accuracy (\%) over FMNIST}
    \begin{tabular}{|c|c|c|c|c|}
    \hline
        Algorithm & Accuracy & Shirt & Pullover & T-shirt \\ \hline
        FedAvg	&85.4	&86.4	&90.1	&79.7\\ \hline
        q-FFL	&84.9	&85.5	&88.6	&80.7\\ \hline
        DR-FedAvg(q=0.1)	&84.7	&86.2	&87.7	&80.7\\ \hline
        DR-FedAvg(q=10)	&84.6	&85.2	&88.9	&79.8\\ \hline
        PGF-FedAvg	&85.2	&87.2	&85.5	&82.6\\ \hline
        DQFFL	&85.3	&87.8	&89.1	&80.7\\
    \hline
    \end{tabular}
\end{table}

\begin{figure}
  \centering 
  \includegraphics[scale=0.35]{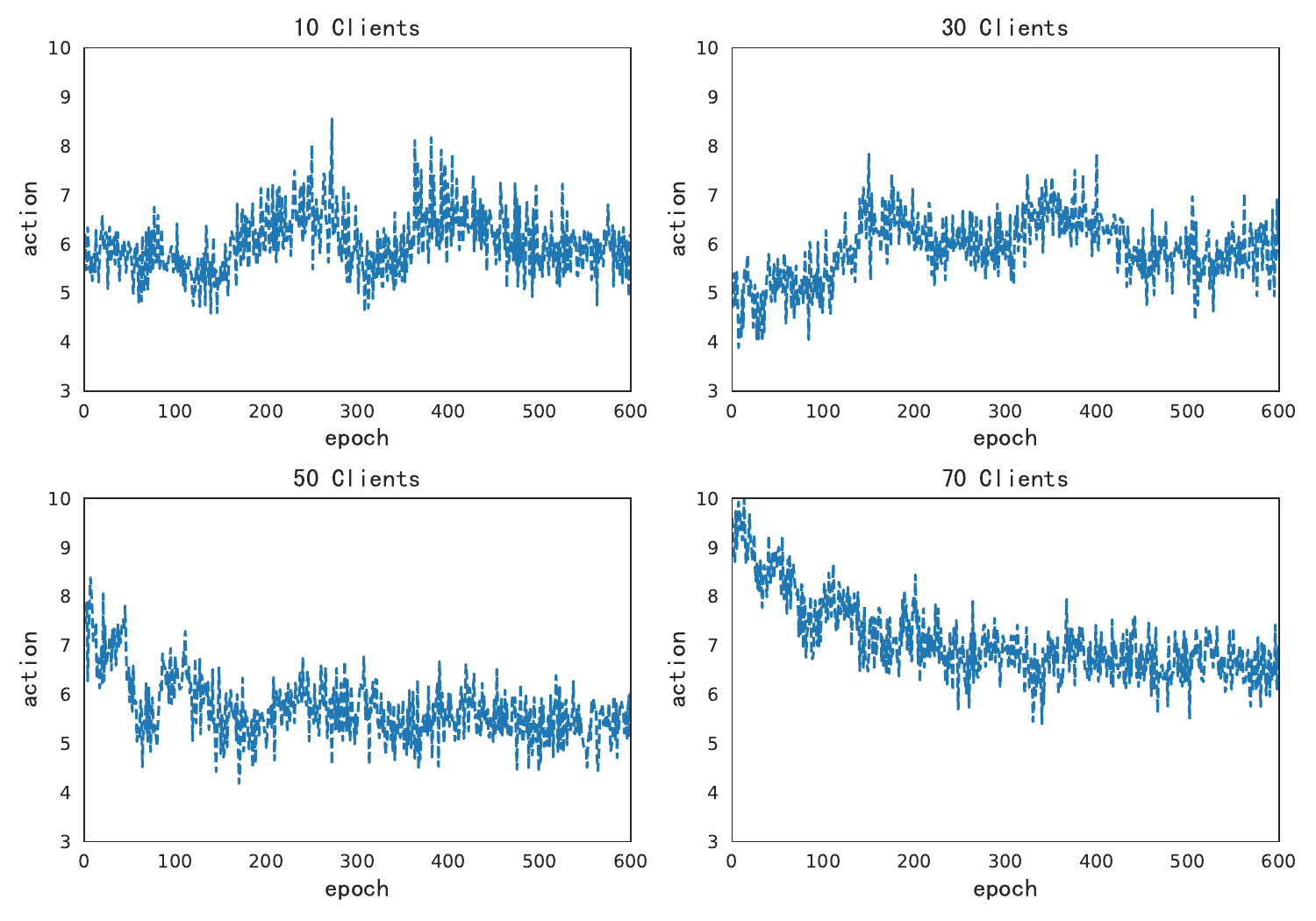}
  \caption{Comparison of Total Clients Number on Action Value of Agents in MNIST Dataset}
  \label{fig3}
\end{figure}

\section{Conclusion}
In this paper, we propose a dynamic fairness-enhanced federated learning algorithm DQFFL, which aims to mitigate client aggregation differences to improve fairer treatment of all groups including minorities in federated learning by converting the maintenance of fairness during federated aggregation to the assignment of client weights to federated aggregation, using the performance of the global federated model on individual clients and incorporating $\alpha$-fairness to achieve the measure of fairness

\ifCLASSOPTIONcompsoc
\else
\fi

\ifCLASSOPTIONcaptionsoff
  \newpage
\fi



%
\bibliographystyle{IEEEtran}
\normalem
\bibliography{research.bib}

\begin{thebibliography}{10}
\providecommand{\url}[1]{#1}
\csname url@samestyle\endcsname
\providecommand{\newblock}{\relax}
\providecommand{\bibinfo}[2]{#2}
\providecommand{\BIBentrySTDinterwordspacing}{\spaceskip=0pt\relax}
\providecommand{\BIBentryALTinterwordstretchfactor}{4}
\providecommand{\BIBentryALTinterwordspacing}{\spaceskip=\fontdimen2\font plus
\BIBentryALTinterwordstretchfactor\fontdimen3\font minus \fontdimen4\font\relax}
\providecommand{\BIBforeignlanguage}[2]{{%
\expandafter\ifx\csname l@#1\endcsname\relax
\typeout{** WARNING: IEEEtran.bst: No hyphenation pattern has been}%
\typeout{** loaded for the language `#1'. Using the pattern for}%
\typeout{** the default language instead.}%
\else
\language=\csname l@#1\endcsname
\fi
#2}}
\providecommand{\BIBdecl}{\relax}
\BIBdecl

\bibitem{li2022federated}
Y.~Li, W.~Li, and Z.~Xue, ``Federated learning with stochastic quantization,'' \emph{International Journal of Intelligent Systems}, vol.~37, no.~12, pp. 11\,600--11\,621, 2022.

\bibitem{zang2023fedpcf}
Y.~Zang, Z.~Xue, S.~Ou, Y.~Long, H.~Zhou, and J.~Du, ``Fedpcf: An integrated federated learning framework with multi-level prospective correction factor,'' in \emph{Proceedings of the 2023 ACM International Conference on Multimedia Retrieval}, 2023, pp. 490--498.

\bibitem{guan2021federated}
Z.~Guan, Y.~Li, Z.~Xue, Y.~Liu, H.~Gao, and Y.~Shao, ``Federated graph neural network for cross-graph node classification,'' in \emph{2021 IEEE 7th International Conference on Cloud Computing and Intelligent Systems (CCIS)}.\hskip 1em plus 0.5em minus 0.4em\relax IEEE, 2021, pp. 418--422.

\bibitem{kou2018hashtag}
F.~Kou, J.~Du, C.~Yang, Y.~Shi, W.~Cui, M.~Liang, and Y.~Geng, ``Hashtag recommendation based on multi-features of microblogs,'' \emph{Journal of Computer Science and Technology}, vol.~33, pp. 711--726, 2018.

\bibitem{meng2013tracking}
D.~Meng, Y.~Jia, J.~Du, and F.~Yu, ``Tracking algorithms for multiagent systems,'' \emph{IEEE Transactions on neural networks and learning systems}, vol.~24, no.~10, pp. 1660--1676, 2013.

\bibitem{mcmahan2017communication}
B.~McMahan, E.~Moore, D.~Ramage, S.~Hampson, and B.~A. y~Arcas, ``Communication-efficient learning of deep networks from decentralized data,'' in \emph{Artificial intelligence and statistics}.\hskip 1em plus 0.5em minus 0.4em\relax PMLR, 2017, pp. 1273--1282.

\bibitem{Jia2022afair}
J.~Tian, X.~Lü, and R.~Zou, ``A fair resource allocation scheme in federated learning,'' \emph{Journal of Computer Research and Development}, 2022.

\bibitem{mo2022multi}
H.~Mo, H.~Zheng, M.~Gao, and X.~Feng, ``Multi-source heterogeneous data fusion based on federated learning,'' \emph{Journal of Computer Research and Development 59.2}, 2022.

\bibitem{bonawitz2019towards}
K.~Bonawitz, H.~Eichner, W.~Grieskamp, D.~Huba, A.~Ingerman, V.~Ivanov, C.~Kiddon, J.~Kone{\v{c}}n{\`y}, S.~Mazzocchi, B.~McMahan \emph{et~al.}, ``Towards federated learning at scale: System design,'' \emph{Proceedings of machine learning and systems}, vol.~1, pp. 374--388, 2019.

\bibitem{wang2021novel}
J.~Wang, Q.~Liu, H.~Liang, G.~Joshi, and H.~V. Poor, ``A novel framework for the analysis and design of heterogeneous federated learning,'' \emph{IEEE Transactions on Signal Processing}, vol.~69, pp. 5234--5249, 2021.

\bibitem{horvath2021fjord}
S.~Horvath, S.~Laskaridis, M.~Almeida, I.~Leontiadis, S.~Venieris, and N.~Lane, ``Fjord: Fair and accurate federated learning under heterogeneous targets with ordered dropout,'' \emph{Advances in Neural Information Processing Systems}, vol.~34, pp. 12\,876--12\,889, 2021.

\bibitem{li2020federated}
T.~Li, A.~K. Sahu, M.~Zaheer, M.~Sanjabi, A.~Talwalkar, and V.~Smith, ``Federated optimization in heterogeneous networks,'' \emph{Proceedings of Machine learning and systems}, vol.~2, pp. 429--450, 2020.

\bibitem{lee2022privacy}
J.-W. Lee, H.~Kang, Y.~Lee, W.~Choi, J.~Eom, M.~Deryabin, E.~Lee, J.~Lee, D.~Yoo, Y.-S. Kim \emph{et~al.}, ``Privacy-preserving machine learning with fully homomorphic encryption for deep neural network,'' \emph{IEEE Access}, vol.~10, pp. 30\,039--30\,054, 2022.

\bibitem{sun2022fair}
Y.~Sun, S.~Si, J.~Wang, Y.~Dong, Z.~Zhu, and J.~Xiao, ``A fair federated learning framework with reinforcement learning,'' in \emph{2022 International Joint Conference on Neural Networks (IJCNN)}.\hskip 1em plus 0.5em minus 0.4em\relax IEEE, 2022, pp. 1--8.

\bibitem{jung2019fairness}
D.-H. Jung, M.-S. Shin, and J.-G. Ryu, ``Fairness-based superframe design and resource allocation for dynamic rate adaptation in dvb-rcs2 satellite systems,'' \emph{IEEE communications letters}, vol.~23, no.~11, pp. 2046--2049, 2019.

\bibitem{li2021ditto}
T.~Li, S.~Hu, A.~Beirami, and V.~Smith, ``Ditto: Fair and robust federated learning through personalization,'' in \emph{International Conference on Machine Learning}.\hskip 1em plus 0.5em minus 0.4em\relax PMLR, 2021, pp. 6357--6368.

\bibitem{sultana2022eiffel}
A.~Sultana, M.~M. Haque, L.~Chen, F.~Xu, and X.~Yuan, ``Eiffel: Efficient and fair scheduling in adaptive federated learning,'' \emph{IEEE Transactions on Parallel and Distributed Systems}, vol.~33, no.~12, pp. 4282--4294, 2022.

\bibitem{mitra2021linear}
A.~Mitra, R.~Jaafar, G.~J. Pappas, and H.~Hassani, ``Linear convergence in federated learning: Tackling client heterogeneity and sparse gradients,'' \emph{Advances in Neural Information Processing Systems}, vol.~34, pp. 14\,606--14\,619, 2021.

\bibitem{wang2021federated}
Z.~Wang, X.~Fan, J.~Qi, C.~Wen, C.~Wang, and R.~Yu, ``Federated learning with fair averaging,'' \emph{arXiv e-prints}, pp. arXiv--2104, 2021.

\bibitem{cotter2019optimization}
A.~Cotter, H.~Jiang, M.~R. Gupta, S.~L. Wang, T.~Narayan, S.~You, and K.~Sridharan, ``Optimization with non-differentiable constraints with applications to fairness, recall, churn, and other goals.'' \emph{J. Mach. Learn. Res.}, vol.~20, no. 172, pp. 1--59, 2019.

\bibitem{li2019fair}
T.~Li, M.~Sanjabi, A.~Beirami, and V.~Smith, ``Fair resource allocation in federated learning,'' in \emph{International Conference on Learning Representations}, 2019.

\bibitem{zhang2021deep}
P.~Zhang, C.~Wang, C.~Jiang, and Z.~Han, ``Deep reinforcement learning assisted federated learning algorithm for data management of iiot,'' \emph{IEEE Transactions on Industrial Informatics}, vol.~17, no.~12, pp. 8475--8484, 2021.

\bibitem{zhao2022dynamic}
Z.~Zhao and G.~Joshi, ``A dynamic reweighting strategy for fair federated learning,'' in \emph{ICASSP 2022-2022 IEEE International Conference on Acoustics, Speech and Signal Processing (ICASSP)}.\hskip 1em plus 0.5em minus 0.4em\relax IEEE, 2022, pp. 8772--8776.

\bibitem{yang2021characterizing}
C.~Yang, Q.~Wang, M.~Xu, Z.~Chen, K.~Bian, Y.~Liu, and X.~Liu, ``Characterizing impacts of heterogeneity in federated learning upon large-scale smartphone data,'' in \emph{Proceedings of the Web Conference 2021}, 2021, pp. 935--946.

\bibitem{blum2021one}
A.~Blum, N.~Haghtalab, R.~L. Phillips, and H.~Shao, ``One for one, or all for all: Equilibria and optimality of collaboration in federated learning,'' in \emph{International Conference on Machine Learning}.\hskip 1em plus 0.5em minus 0.4em\relax PMLR, 2021, pp. 1005--1014.

\bibitem{agarwal2018reductions}
A.~Agarwal, A.~Beygelzimer, M.~Dud{\'\i}k, J.~Langford, and H.~Wallach, ``A reductions approach to fair classification,'' in \emph{International conference on machine learning}.\hskip 1em plus 0.5em minus 0.4em\relax PMLR, 2018, pp. 60--69.

\bibitem{spiridonoff2021communication}
A.~Spiridonoff, A.~Olshevsky, and Y.~Paschalidis, ``Communication-efficient sgd: From local sgd to one-shot averaging,'' \emph{Advances in Neural Information Processing Systems}, vol.~34, pp. 24\,313--24\,326, 2021.

\bibitem{kim2022adaptive}
S.~Kim, ``Adaptive multiservice resource allocation algorithm with wireless network virtualization,'' \emph{IEEE Access}, vol.~10, pp. 51\,515--51\,524, 2022.

\bibitem{zhang2021adaptive}
H.~Zhang, Z.~Xie, R.~Zarei, T.~Wu, and K.~Chen, ``Adaptive client selection in resource constrained federated learning systems: A deep reinforcement learning approach,'' \emph{IEEE Access}, vol.~9, pp. 98\,423--98\,432, 2021.

\bibitem{shamir2014communication}
O.~Shamir, N.~Srebro, and T.~Zhang, ``Communication-efficient distributed optimization using an approximate newton-type method,'' in \emph{International conference on machine learning}.\hskip 1em plus 0.5em minus 0.4em\relax PMLR, 2014, pp. 1000--1008.

\bibitem{smith2017federated}
V.~Smith, C.-K. Chiang, M.~Sanjabi, and A.~S. Talwalkar, ``Federated multi-task learning,'' \emph{Advances in neural information processing systems}, vol.~30, 2017.

\bibitem{go2009twitter}
A.~Go, R.~Bhayani, and L.~Huang, ``Twitter sentiment classification using distant supervision,'' \emph{CS224N project report, Stanford}, vol.~1, no.~12, p. 2009, 2009.

\bibitem{zhou2022chinese}
Y.~Zhou, J.~Du, Z.~Xue, A.~Li, and Z.~Guan, ``Chinese word sense embedding with sememewsd and synonym set,'' in \emph{CAAI International Conference on Artificial Intelligence}.\hskip 1em plus 0.5em minus 0.4em\relax Springer, 2022, pp. 236--247.

\bibitem{wei2019boosting}
X.~Wei, J.~Du, M.~Liang, and L.~Ye, ``Boosting deep attribute learning via support vector regression for fast moving crowd counting,'' \emph{Pattern Recognition Letters}, vol. 119, pp. 12--23, 2019.

\bibitem{li2022scientific}
A.~Li, J.~Du, F.~Kou, Z.~Xue, X.~Xu, M.~Xu, and Y.~Jiang, ``Scientific and technological information oriented semantics-adversarial and media-adversarial cross-media retrieval,'' \emph{arXiv preprint arXiv:2203.08615}, 2022.

\bibitem{li2017tobit}
W.~Li, Y.~Jia, and J.~Du, ``Tobit kalman filter with time-correlated multiplicative measurement noise,'' \emph{IET Control Theory \& Applications}, vol.~11, no.~1, pp. 122--128, 2017.

\bibitem{li2017variance}
W.~Li, J.~Sun, Y.~Jia, J.~Du, and X.~Fu, ``Variance-constrained state estimation for nonlinear complex networks with uncertain coupling strength,'' \emph{Digital Signal Processing}, vol.~67, pp. 107--115, 2017.

\bibitem{meng2015robust}
D.~Meng, Y.~Jia, and J.~Du, ``Robust iterative learning protocols for finite-time consensus of multi-agent systems with interval uncertain topologies,'' \emph{International Journal of Systems Science}, vol.~46, no.~5, pp. 857--871, 2015.

\bibitem{cao2013review}
J.~Cao, D.-h. Mao, Q.~Cai, H.-s. Li, and J.-p. Du, ``A review of object representation based on local features,'' \emph{Journal of Zhejiang University SCIENCE C}, vol.~14, no.~7, pp. 495--504, 2013.

\bibitem{meng2015high}
D.~Meng, Y.~Jia, J.~Du, and J.~Zhang, ``High-precision formation control of nonlinear multi-agent systems with switching topologies: A learning approach,'' \emph{International journal of robust and nonlinear control}, vol.~25, no.~13, pp. 1993--2018, 2015.

\bibitem{li2018resilient}
W.~Li, Y.~Jia, and J.~Du, ``Resilient filtering for nonlinear complex networks with multiplicative noise,'' \emph{IEEE Transactions on Automatic Control}, vol.~64, no.~6, pp. 2522--2528, 2018.

\bibitem{li2023multi}
A.~Li, Y.~Li, Y.~Shao, and B.~Liu, ``Multi-view scholar clustering with dynamic interest tracking,'' \emph{IEEE Transactions on Knowledge and Data Engineering}, 2023.

\bibitem{li2022predicting}
Y.~Li, I.~Y. Zeng, Z.~Niu, J.~Shi, Z.~Wang, and Z.~Guan, ``Predicting vehicle fuel consumption based on multi-view deep neural network,'' \emph{Neurocomputing}, vol. 502, pp. 140--147, 2022.

\bibitem{li2021heterogeneous}
Y.~Li, D.~Jiang, R.~Lian, X.~Wu, C.~Tan, Y.~Xu, and Z.~Su, ``Heterogeneous latent topic discovery for semantic text mining,'' \emph{IEEE Transactions on Knowledge and Data Engineering}, vol.~35, no.~1, pp. 533--544, 2021.

\bibitem{xiao2022lecf}
S.~Xiao, Y.~Shao, Y.~Li, H.~Yin, Y.~Shen, and B.~Cui, ``Lecf: recommendation via learnable edge collaborative filtering,'' \emph{Science China Information Sciences}, vol.~65, no.~1, p. 112101, 2022.

\bibitem{shao2021memory}
Y.~Shao, S.~Huang, Y.~Li, X.~Miao, B.~Cui, and L.~Chen, ``Memory-aware framework for fast and scalable second-order random walk over billion-edge natural graphs,'' \emph{The VLDB Journal}, vol.~30, no.~5, pp. 769--797, 2021.

\bibitem{li2020distributed}
Y.~Li, Y.~Yuan, Y.~Wang, X.~Lian, Y.~Ma, and G.~Wang, ``Distributed multimodal path queries,'' \emph{IEEE Transactions on Knowledge and Data Engineering}, vol.~34, no.~7, pp. 3196--3210, 2020.

\bibitem{li2019application}
Y.~Li, L.~Yang, B.~Yang, N.~Wang, and T.~Wu, ``Application of interpretable machine learning models for the intelligent decision,'' \emph{Neurocomputing}, vol. 333, pp. 273--283, 2019.

\bibitem{li2018neural}
Y.~Li, W.~Jiang, L.~Yang, and T.~Wu, ``On neural networks and learning systems for business computing,'' \emph{Neurocomputing}, vol. 275, pp. 1150--1159, 2018.

\end{thebibliography}

%

\end{document}